%% file: main.tex
\documentclass{article}

\usepackage{microtype}
\usepackage{graphicx}
\usepackage{subfigure}
\usepackage{booktabs} 

\usepackage{hyperref}

\usepackage[accepted]{icml2024}

\usepackage{amsmath}
\usepackage{amssymb}
\usepackage{amsthm}
\usepackage{amsfonts}
\usepackage{caption}

\usepackage{listings}
\usepackage{xcolor}
\definecolor{darkgreen}{rgb}{0,0.6,0}

\lstdefinestyle{python}{
    language=Python,
    basicstyle=\footnotesize\ttfamily,
    keywordstyle=\color{blue},
    stringstyle=\color{red},
    commentstyle=\color{darkgreen},
    showstringspaces=false,
    breaklines=true,
}

\icmltitlerunning{Scaling Down Deep Learning with MNIST-1D}

\begin{document}

\twocolumn[
\icmltitle{Scaling Down Deep Learning with MNIST-1D}


\begin{icmlauthorlist}
\icmlauthor{Sam Greydanus}{osu,ml}
\icmlauthor{Dmitry Kobak}{tueb,heid}
\end{icmlauthorlist}

\icmlaffiliation{osu}{Oregon State University, USA}
\icmlaffiliation{ml}{The ML Collective}
\icmlaffiliation{tueb}{University of T\"ubingen, Germany}
\icmlaffiliation{heid}{Heidelberg University, Germany}
\icmlcorrespondingauthor{Sam Greydanus}{samgreydanus@gmail.com}

\icmlkeywords{MNIST, dataset, benchmark, deep learning, convolution}

\vskip 0.3in
]



\printAffiliationsAndNotice{}  

\begin{abstract}
Although deep learning models have taken on commercial and political relevance, key aspects of their training and operation remain poorly understood. This has sparked interest in \textit{science of deep learning} projects, many of which require large amounts of time, money, and electricity. But how much of this research really needs to occur at scale? In this paper, we introduce MNIST-1D: a minimalist, procedurally generated, low-memory, and low-compute alternative to classic deep learning benchmarks. Although the dimensionality of MNIST-1D is only 40 and its default training set size only 4000, MNIST-1D can be used to study inductive biases of different deep architectures, find lottery tickets, observe deep double descent, metalearn an activation function, and demonstrate guillotine regularization in self-supervised learning. All these experiments can be conducted on a GPU or often even on a CPU \textit{within minutes}, allowing for fast prototyping, educational use cases, and cutting-edge research on a low budget.
\end{abstract}

\input{content}

\section*{Acknowledgements}

We would like to thank Luke Metz for the interesting conversations, Tony Zador for the encouragement to release this dataset, Simon Prince for improving our initial double descent experiments and for the permission to adapt his code for Figure \ref{fig:deep_double_descent}, and Peter Steinbach for his help with preparing the Python package.

This work was partially funded by Deutsche Forschungsgemeinschaft (DFG, German Research Foundation) via Germany’s Excellence Strategy (Excellence cluster 2064 ``Machine Learning --- New Perspectives for Science'', EXC 390727645; Excellence cluster 2181 ``STRUCTURES'', EXC 390900948), the German Ministry of Science and Education (BMBF) via the T\"{u}bingen AI Center (01IS18039A), and the Gemeinn\"{u}tzige Hertie-Stiftung.

\section*{Impact Statement}

The goal of our paper is to advance the field of machine learning. We do not see any potential societal consequences of our work that need to be highlighted in this section.

\bibliographystyle{abbrvnat}
\setlength{\bibsep}{5pt} 
\setlength{\bibhang}{0pt}
\bibliography{references}

\clearpage

\renewcommand{\thefigure}{S\arabic{figure}}
\setcounter{figure}{0}  
\renewcommand{\thetable}{S\arabic{table}}
\setcounter{table}{0} 
\renewcommand{\theHtable}{Supplement.\thetable}
\renewcommand{\theHfigure}{Supplement.\thefigure}

\appendix

\input{appendix}

\end{document}

%% file: content.tex
\section{Introduction}

The deep learning analogue of \textit{Drosophila melanogaster} is the MNIST dataset. Drosophila, the fruit fly, has a life cycle that is just a few days long, its nutritional needs are negligible, and it is easier to work with than mammals, especially humans. Like Drosophila, MNIST is easy to use: training a classifier on it takes only a few a minutes whereas training full-size vision and language models can take months of time and millions of dollars \cite{sharir2020cost}.

\begin{figure}[t]
    \includegraphics[width=\columnwidth]{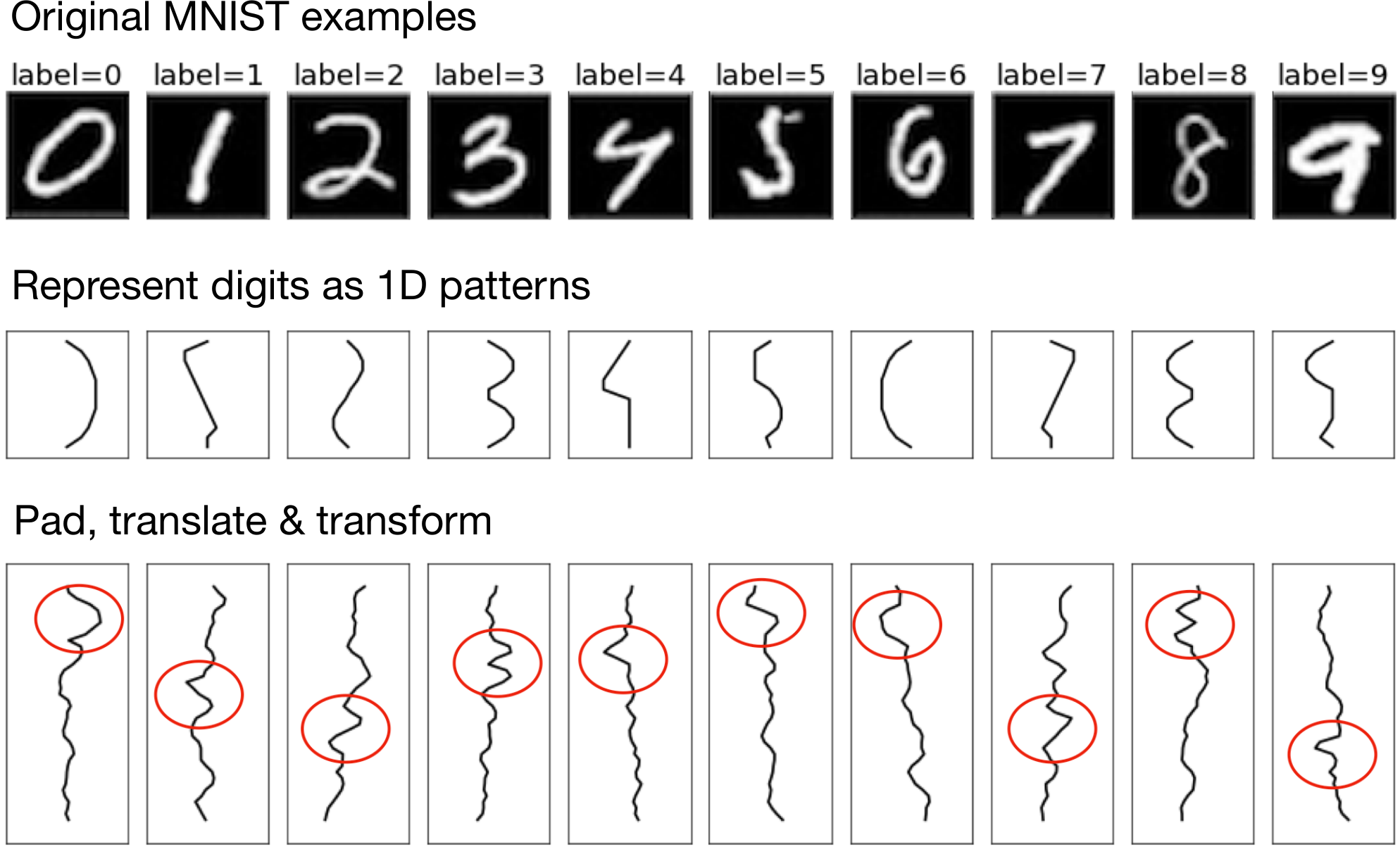}
    \caption{Constructing the MNIST-1D dataset. Unlike MNIST, each sample is a one-dimensional sequence. To generate each sample, we begin with a hand-crafted digit template loosely inspired by MNIST shapes. Then we randomly pad, translate, and add noise to produce 1D sequences with 40 points each. \hfill\href{https://github.com/greydanus/mnist1d/blob/master/notebooks/building-mnist1d.ipynb}{\texttt{\textbf{[CODE]}}} }
    \label{fig:overview}
\end{figure}

But in spite of their small size, both test systems have had a major impact on their respective fields. A number of seminal discoveries in medicine, including multiple Nobel prizes, have been awarded for work performed with fruit flies. Early work in genetics --- work that paved the way for the Human Genome Project, which involved billions of dollars of funding, dozens of institutions, and over a decade of accelerated research \cite{lander2001initial} --- was performed on fruit flies and other simple organisms. To this day,  experiments on Drosophila are a cornerstone of biomedical research.

\begin{table*}[t]
\caption{Test accuracies (in \%) of common classifiers on the MNIST and MNIST-1D datasets. Most classifiers achieve similar test accuracy on MNIST. By contrast, the MNIST-1D dataset is capable of separating different models based on their inductive biases. The drop in CNN and GRU performance when using shuffled features indicates that spatial priors are important on this dataset. All models except logistic regression achieve 100\% accuracy on the training set. Standard deviation is computed over three runs.\hfill\href{https://github.com/greydanus/mnist1d/blob/master/notebooks/mnist1d-classification.ipynb}{\texttt{\textbf{[CODE]}}}}
\label{tab:main-benchmark}
\begin{center} \begin{small}
\begin{tabular}{lccccc}
\toprule
Dataset	& Logistic regression & MLP & CNN & GRU & Human expert \\
\midrule
MNIST    &  $94 \pm 0.5$ &  $>99$ &  $>99$ & $>99$ &  $>99$ \\
MNIST-1D   & $32 \pm 1$ &  $68 \pm 2$ &  $94 \pm 2$ &  $91 \pm 2$&  $96 \pm 1$  \\
MNIST-1D (shuffled)   & $32 \pm 1$ &  $68 \pm 2$ &  $56 \pm 2$ &  $57 \pm 2$ &  $\sim 30 \pm 10$  \\
\bottomrule
\end{tabular}
\end{small} \end{center}
\vskip -0.1in
\end{table*}

Meanwhile, MNIST has served as the initial proving ground for a large number of deep learning innovations including dropout, Adam, convolutional networks, generative adversarial networks, and variational autoencoders \cite{srivastava2014dropout, kingma2014adam, lecun1989backpropagation, goodfellow2014generative, kingma2013auto}. Once a proof of concept was established in small-scale experiments, researchers were able to justify the time and resources needed for larger and more impactful applications.

However, despite its historical significance, MNIST has three notable shortcomings. First, it is too simple. Linear classifiers, fully-connected networks, and convolutional models all perform similarly well, obtaining above 90\% accuracy (Table~\ref{tab:main-benchmark}). This makes it hard to measure the contribution of a CNN's spatial priors or to judge the relative effectiveness of different regularization schemes. Second, it is too large. Each sample in MNIST is a $28\times 28$ image, resulting in 784 input dimensions. Together with its sample size $n=70\,000$, this requires an unnecessarily large amount of computation to perform a hyperparameter search or debug a metalearning loop. Third, it is hard to hack. MNIST is a fixed dataset and it is difficult to increase the sample size or to change the noise distribution. The ideal toy dataset should be procedurally generated to allow researchers to vary its parameters at will.

In order to address these shortcomings, we propose the MNIST-1D dataset (Figure~\ref{fig:overview}). It is a minimalist, low-memory, and low-compute alternative to MNIST, designed for exploratory deep learning research where rapid prototyping and short latency are a priority. MNIST-1D has 40 dimensions, many fewer than MNIST's 784 or CIFAR's 3,072. The sample size can be arbitrarily large, but the frozen default dataset contains 4000 training and 1000 test samples, many fewer than the 70,000 in MNIST and 60,000 in CIFAR-10/100. Although our dataset is procedurally generated, its samples are intuitive enough for a human expert to match or even  outperform a CNN. 

MNIST-1D does a much better job than the original MNIST at differentiating between model architectures: a linear classifier can only achieve 32\% accuracy (Table~\ref{tab:main-benchmark}), while a CNN reaches 94\%. Below we show that MNIST-1D can be used to study phenomena ranging from deep double descent to self-supervised learning. Crucially, the experiments we present in this paper take only a few minutes to run on a single GPU (in some cases just a CPU) whereas they would require multiple GPU hours or even GPU days when using MNIST or CIFAR. This makes MNIST-1D valuable as a playground for quick initial experiments and invaluable for researchers without access to powerful GPUs.

All our experiments are in Jupyter notebooks and are available at \url{https://github.com/greydanus/mnist1d}, with direct links from figure captions. We provide a \texttt{mnist1d} package that can be installed via \texttt{pip install mnist1d}.

\section{Related work}

There are a number of small-scale datasets that are commonly used to investigate \textit{science of deep learning} questions. We have already alluded to MNIST \cite{lecun-1998-IEEE-gradient-based-learning-applied}, CIFAR-10 and CIFAR-100 \citep{krizhevsky2009learning}. The CIFAR datasets consist of $32\times 32$ colored natural images and have sample sizes similar to MNIST. They are better at discriminating between MLP and CNN architectures and also between different types of CNNs: for example, vanilla CNNs versus ResNets \cite{he-2015-arXiv-deep-residual-learning}. The FashionMNIST dataset \citep{xiao2017} has the same size as MNIST and is somewhat more difficult. It aims to rectify some of the most serious problems with MNIST: in particular, that MNIST is too easy and thus all neural network models attain roughly the same test accuracy. None of these datasets is substantially smaller than MNIST and this hampers their use in fast-paced exploratory research or compute-heavy applications such as metalearning.

There are very few datasets smaller than MNIST that are of interest for deep learning research. Toy datasets provided by Scikit-learn \cite{scikit-learn}, such as the \texttt{two\_moons} dataset, can be useful for studying clustering or training very simple classifiers, but are not sufficiently complex for deep learning investigations. Indeed, these datasets are just 2D point clouds, devoid of spatial or temporal correlations between features and lacking manifold structures that a deep nonlinear classifier could use to escape the curse of dimensionality \cite{bellman1959adaptive}.

To the best of our knowledge, the MNIST-1D dataset is unique in that it is over two orders of magnitude smaller than MNIST but can be used just as effectively --- and in a number of important cases, \textit{more effectively} --- for studying fundamental deep learning questions. This may be why MNIST-1D was used as a teaching tool in the recent \textit{Understanding Deep Learning} textbook \cite{prince2023understanding}\footnote{The initial preprint of this manuscript was released on arXiv in November 2020.}.

MNIST-1D bears philosophical similarities to the \textit{Synthetic Petri Dish} by \citet{rawal2020synthetic}. The authors make similar references to biology in order to motivate the use of small synthetic datasets for exploratory research. Their work differs from ours in that they use metalearning to obtain their datasets whereas we construct ours by hand. In doing so, we are able to control various causal factors such as the amount of noise, translation, and padding. Our dataset is more intuitive to humans: an experienced human can outperform a strong CNN on the MNIST-1D classification task. This is not possible on the Synthetic Petri Dish dataset.

\section{The MNIST-1D dataset}

\begin{table}[t]
\caption{Default parameters for MNIST-1D generation.}
\label{tab:inner-hypers}
\begin{center}
\begin{tabular}{ll}
\toprule
\textbf{Parameter} & \textbf{Value} \\
\midrule
Train/test split    &  4000/1000  \\
Template length    &  12  \\
Padding points    &  36--60  \\
Max. translation    &  48  \\
Gaussian filter width    &  2  \\
Gaussian noise scale    &  0.25  \\
White noise scale    &  0.02  \\
Shear scale    &  0.75  \\
Final seq. length    &  40  \\
Random seed    &  42  \\
\bottomrule
\end{tabular}
\end{center}
\end{table}

\textbf{Dimensionality.} Our first design choice was to use one-dimensional time series instead of two-dimensional grayscale images or three-dimensional tensors corresponding to colored images. Our rationale was that one-dimensional signals require far less computation to train on but can be designed to have many of the same biases, distortions, and distribution shifts that are of interest to researchers studying fundamental deep learning questions.

\textbf{Constructing the dataset.} We began with ten one-dimensional template patterns which resemble the digits 0--9 when plotted as in Figure \ref{fig:overview}. Each of these templates consisted of 12 hand-crafted $x$ coordinates. Next we padded the end of each sequence with 36--60 additional zero values, did a random circular shift by up to 48 indices, applied a random scaling, added Gaussian noise, and added a constant linear signal. We used Gaussian smoothing with $\sigma=2$ to induce spatial correlations. Finally, we downsampled the sequences to 40 data points that play the role of \textit{pixels} in the resulting MNIST-1D (Figure~\ref{fig:overview}). Table~\ref{tab:inner-hypers} gives the values of all the default hyperparameters used in these transformations.

\textbf{Implementation.} Our goal was to make the code as simple, modular, and extensible as possible. The code for generating the dataset occupies two Python files and fits in a total of 150 lines. The \texttt{get\_dataset} method has a simple API for changing dataset features such as maximum digit translation, correlated noise scale, shear scale, final sequence length, and more (Table~\ref{tab:inner-hypers}). The following code snippet shows how to install the \texttt{mnist1d} package, choose a custom number of samples, and generate a dataset:




\begin{lstlisting}
# install the package from PyPI
# pip install mnist1d

from mnist1d.data import make_dataset
from mnist1d.data import get_dataset_args

args = get_dataset_args() # default params
args.num_samples = 10_000
data = make_dataset(args)
x, y = data["x"], data["y"]
\end{lstlisting}

The frozen dataset with $4000 + 1000$ samples can be found on GitHub as \texttt{mnist1d\_data.pkl}.

\section{Classification}

\begin{figure}[t]
    \includegraphics[width=\columnwidth]{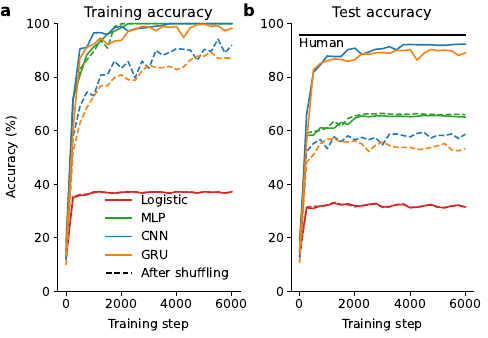}
    \caption{Train and test accuracy of common classification models on MNIST-1D. The logistic regression model fares worse than the MLP. Meanwhile, the MLP fares worse than the CNN and GRU, which use translation invariance and local connectivity to bias optimization towards solutions that generalize well. When local spatial correlations are destroyed by shuffling feature indices (dashed lines), the MLP performs the best. CPU runtime: $\sim$10 minutes.\hfill\href{https://github.com/greydanus/mnist1d/blob/master/notebooks/mnist1d-classification.ipynb}{\texttt{\textbf{[CODE]}}}}
    \label{fig:benchmarks}
\end{figure}

We used PyTorch to implement and train simple logistic, MLP (fully-connected), CNN (with 1D convolutions), and GRU (gated recurrent unit) models. We used the Adam optimizer and early stopping for model selection and evaluation. We obtained 32\% accuracy with logistic regression, 68\% using an MLP, 91\% using a GRU, and 94\% using a CNN (Table~\ref{tab:main-benchmark}). Even though the test performance was markedly different between MLP, GRU, and CNN, all of them easily achieved 100\% accuracy on the training set (Figure~\ref{fig:benchmarks}). While for the MLP this is a manifestation of overfitting, for the CNN this is an example of \textit{benign overfitting}.

For comparison, we also report the accuracy of a human expert (one of the authors) trained on the training set and evaluated (one-shot) on the test set. His accuracy was 96\%. The purpose of this comparison was to show that MNIST-1D is a task that is as intuitive for humans as it is for machine learning models with spatial priors. This suggests that the models are not achieving high performance by exploiting some unintuitive statistical artifacts. Rather, they are using the relative position of various features associated with each 1-D digit. Interestingly, the CNN and the human expert had similar per-digit error rates (Figure~\ref{fig:distribution}).

\paragraph{Shuffling sanity check.} We also trained the same models on a version of the dataset which was permuted along the spatial dimension. This `shuffled' version measured each of the models' performances in the absence of local spatial structure \citep{zhang-2017-ICLR-understanding-deep-learning, li-2018-ICLR-measuring-the-intrinsic-dimension}. The test accuracy of CNNs and GRUs decreased by about 35 percentage points after shuffling whereas the MLP and logistic models performed about the same (Table~\ref{tab:main-benchmark}, Figure~\ref{fig:benchmarks}). This makes sense, as the former two models have spatial and temporal locality priors whereas the latter two do not.

\begin{figure}[t]
    \centering
    \includegraphics[width=\columnwidth]{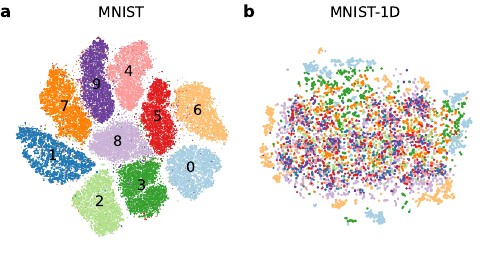}
    \caption{Visualizing the MNIST and MNIST-1D datasets with $t$-SNE. The well-defined clusters in the MNIST embedding indicate that the classes are separable via a simple $k$NN classifier in pixel space. The MNIST-1D plot reveals little structure and a lack of clusters, indicating that nearest neighbors in pixel space are not semantically meaningful, as is the case with natural image datasets.
    \hfill\href{https://github.com/greydanus/mnist1d/blob/master/notebooks/tsne-mnist-vs-mnist1d.ipynb}{\texttt{\textbf{[CODE]}}}}
    \label{fig:tsne}
\end{figure}

\textbf{Dimensionality reduction.} We used $t$-SNE \citep{van-der-maaten2008visualizing-data-using} to visualize MNIST and MNIST-1D in two dimensions. We observed ten well-defined clusters in the MNIST dataset, suggesting that the classes are separable with a $k$NN classifier in pixel space (Figure~\ref{fig:tsne}a). In contrast, there were few well-defined clusters in the MNIST-1D visualization, suggesting that the nearest neighbors in pixel space are not semantically meaningful (Figure~\ref{fig:tsne}b). This is well known to be the case for natural image datasets such as CIFAR-10/100, and is therefore a \textit{benefit} of MNIST-1D, making it more interesting.

\section{Science of deep learning with MNIST-1D}

In this section we show how MNIST-1D can be used to explore empirical science of deep learning topics.

\subsection{Lottery tickets and spatial inductive biases} 

\begin{figure*}[t]
    \includegraphics[width=\textwidth]{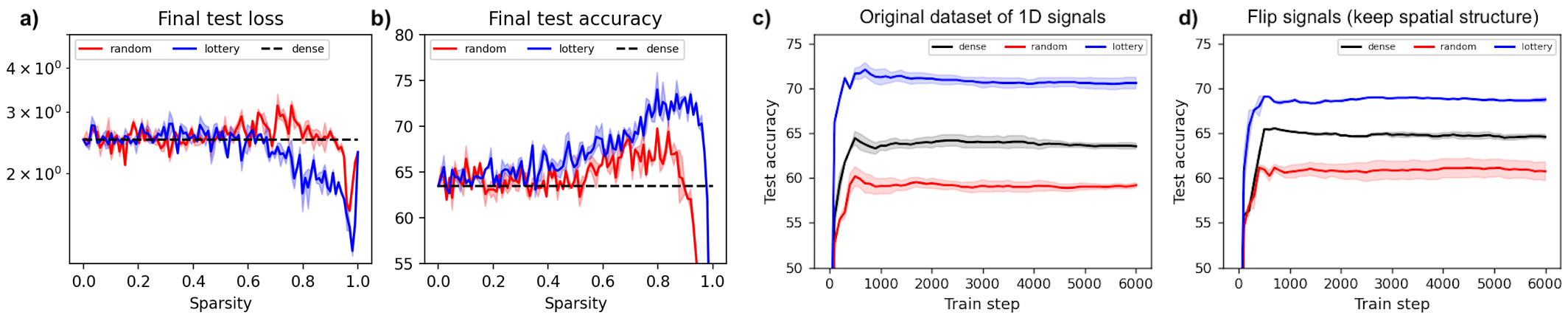}
    \includegraphics[width=\textwidth]{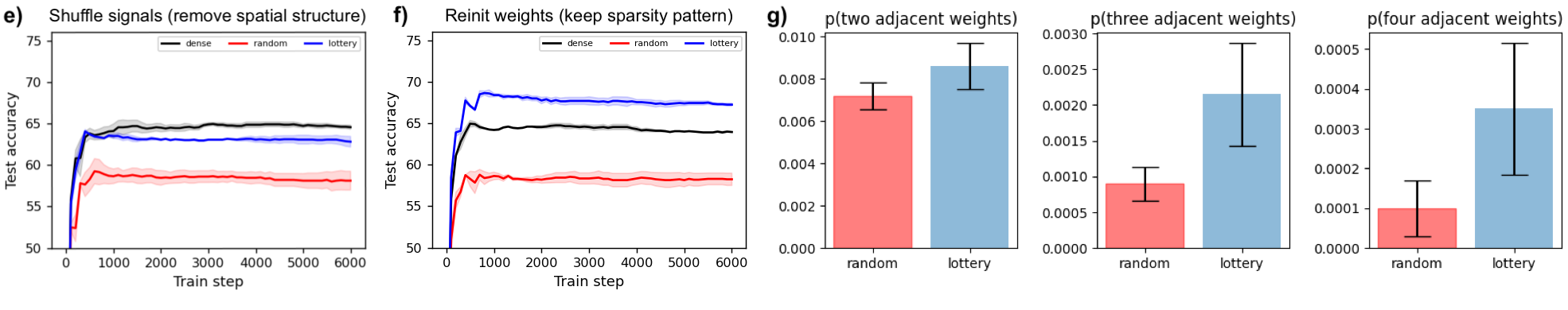}
    \caption{Finding and analyzing lottery tickets. \textbf{(a--b)} The test loss and test accuracy of lottery tickets at different levels of sparsity, compared to randomly selected subnetworks and to the original dense network. \textbf{(c)} Performance of lottery tickets with 92\% sparsity. \textbf{(d)} Performance of the same lottery tickets when trained on flipped data. \textbf{(e)} Performance of the same lottery tickets when trained on data with shuffled features. \textbf{(f)} Performance of the same lottery tickets but with randomly initialized weights, when trained on original data. \textbf{(g)} Lottery tickets had more adjacent non-zero weights in the first layer compared to random subnetworks. Runtime: $\sim$30 minutes. \hfill \href{https://github.com/greydanus/mnist1d/blob/master/notebooks/lottery-tickets.ipynb}{\texttt{\textbf{[CODE]}}} }
    \label{fig:lottery}
\end{figure*}

It is not unusual for deep learning models to have many times more parameters than necessary to perfectly fit the training set \citep{prince2023understanding}. This overparameterization helps training but increases computational overhead. One solution is to progressively prune weights from a model during training so that the final network is just a fraction of its original size. Although this approach works, conventional wisdom holds that sparse networks do not train well from scratch. Recent work by \citet{frankle-2019-ICLR-the-lottery-ticket-hypothesis} challenges this conventional wisdom. The authors report finding sparse subnetworks inside of larger networks that can be trained in isolation to equivalent or even higher accuracies. These \textit{lottery ticket} subnetworks can be found through a simple iterative procedure: train a network, prune the smallest weights, reset the remaining weights to their original values at initialization, and then retrain and repeat the process until the desired sparsity threshold is reached.

Since the original paper was published, many works have sought to explain this phenomenon and then harness it on larger datasets and models. However, very few works have attempted to isolate a minimal working example of this effect so as to investigate it more carefully. We were able to demonstrate the existence of lottery tickets in a MLP classifier trained on MNIST-1D (Figure \ref{fig:lottery}a--b). Lottery ticket subnetworks that we found performed better than random subnetworks with the same level of sparsity. Remarkably, even at high ($>$95\%) rates of sparsity, the lottery tickets we found performed \textit{better} than the original dense network.

The asymptotic performance of lottery tickets with 92\% sparsity was around 70\% (Figure~\ref{fig:lottery}c). When we reversed all the 1D patterns in the dataset, effectively preserving the spatial structure but changing the actual locations of all features (analogous to flipping an image upside down), the original lottery tickets continued to perform at around 70\% accuracy (Figure~\ref{fig:lottery}d). This suggests that the lottery tickets did not overfit to the original dataset; instead, something about their connectivity and initial weights gave them an inherent advantage over random sparse networks. This reproduces the findings of \citet{morcos2019one}, which showed that lottery tickets can transfer between datasets.

Next, we asked whether spatial inductive biases were a factor in the high performance of the lottery tickets we had found. To answer this question, we trained the same tickets on a feature-shuffled version of the MNIST-1D dataset. In other words, we permuted the feature indices in order to remove any spatial structure from the data. Shuffling greatly reduced the performance of the lottery tickets: they performed appreciably worse --- worse, in fact, than the original dense network (Figure~\ref{fig:lottery}e). This suggests that part of the lottery tickets' performance can be attributed to a spatial inductive bias in their sparse connectivity structure.

Furthermore, on the original (non-shuffled) MNIST-1D, when we froze the sparsity patterns of lottery tickets but initialized them with different random weights, they still continued to outperform the original dense network (Figure~\ref{fig:lottery}f). This suggests that not the weight values but rather the sparsity patterns represent the spatial inductive bias of lottery tickets. We verified this hypothesis by measuring how often non-zero weights in a lottery ticket were adjacent to each other in the first layer of the model. The lottery tickets had more adjacent weights than expected by chance (Figure~\ref{fig:lottery}g), implying a bias towards local connectivity. See Figure~\ref{fig:lottery_masks} for a visualization of the actual sparsity patterns of several lottery tickets.

The original lottery ticket paper \citep{frankle-2019-ICLR-the-lottery-ticket-hypothesis}, as well as some of the follow-up studies \citep{morcos2019one}, required  a large number of GPUs and multiple days of runtime. By contrast, all the experiments we presented here took around $\sim$30 minutes to complete on a single GPU.

\subsection{Deep double descent}

\begin{figure}[t]
    \centering
    \includegraphics[width=\columnwidth]
    {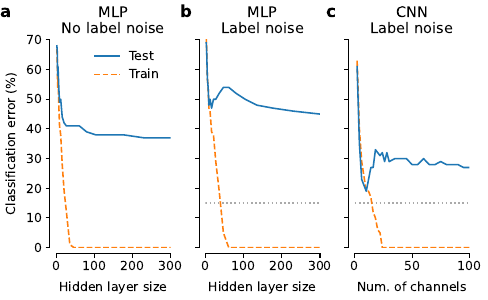}
    \caption{Deep double descent in MNIST-1D classification. Here the test set had $12\,000$ samples. \textbf{(a)} MLP classifier with one hidden layer. \textbf{(b)} MLP classifier; 15\% label noise. \textbf{(c)} CNN classifier with three convolutional layers; 15\% label noise. Adapted with permission from \citet[Section 8.4]{prince2023understanding}. CPU runtime: $\sim$60 minutes. \hfill\href{https://github.com/greydanus/mnist1d/blob/master/notebooks/deep-double-descent.ipynb}{\texttt{\textbf{[CODE]}}}}
    \label{fig:deep_double_descent}
\end{figure}

An intriguing property of neural networks is the \textit{double descent} phenomenon. This phrase refers to a training regime where more data, more model parameters, or more gradient descent steps can \textit{reduce} the test accuracy before it increases again \cite{trunk1979problem,belkin2018reconciling, geiger2019jamming, nakkiran2019deep}. This happens around the so-called interpolation threshold where the learning procedure, consisting of a model and an optimization algorithm, is just barely able to fit the entire training set. At this threshold there is effectively just a single model that can fit the data and this model is very sensitive to label noise and model mis-specification, resulting in overfitting and poor test performance. In contrast, larger models tend to exhibit \textit{benign overfitting} wherein SGD selects a smooth model out of the many possible models fitting the training set (\textit{implicit regularization}). 

Despite the above intuition, many aspects of double descent, such as what factors affect its width and location, are not well understood. We argue that MNIST-1D is well suited for exploring these questions. We observed double descent when training a MLP classifier on MNIST-1D, varying the size of the single hidden layer. In the presence of 15\% label noise, the test error peaked at the interpolation threshold (training error reaching zero), at around 50 neurons in the hidden layer (Figure~\ref{fig:deep_double_descent}b). Further increasing the model size led to the test error dropping again. Without label noise, the test error did not peak (Figure~\ref{fig:deep_double_descent}a). We observed qualitatively similar behavior using the CNN architecture (Figure~\ref{fig:deep_double_descent}c). 
The runtime of this experiment was $\sim$60 minutes on a CPU.

\subsection{Gradient-based metalearning}

The goal of metalearning is to \textit{learn how to learn}. This can be implemented by having two levels of optimization: a fast inner optimization loop which corresponds to a traditional learning objective and a slow outer loop which updates some meta properties of the learning process. One of the simplest examples of metalearning is gradient-based hyperparameter optimization. This concept was proposed in \citet{bengio2000gradient} and then scaled to deep learning models by \citet{maclaurin2015gradient}. The basic idea is to implement a fully differentiable  training loop and then backpropagate through the entire process in order to optimize hyperparameters such as the learning rate or the weight decay.

\begin{figure}[t]
    \centering
    \includegraphics[width=0.8\columnwidth]{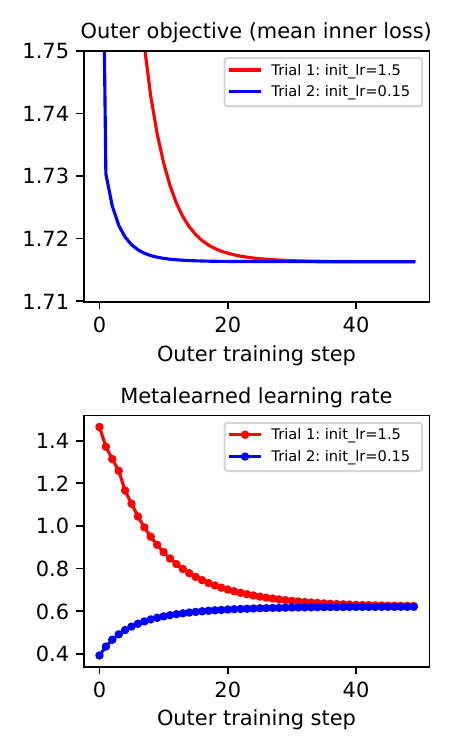}
    \caption{Metalearning the learning rate of SGD optimization of an MLP classifier on MNIST-1D. The outer training converges to the optimal learning rate of 0.62 regardless of whether the initial learning rate is too high or too low. Runtime: $\sim$1 minute. \hfill\href{https://github.com/greydanus/mnist1d/blob/master/notebooks/metalearn-learn-rate.ipynb}{\texttt{\textbf{[CODE]}}} }
    \label{fig:metalearn_lr}
\end{figure}

Metalearning is a promising line of research but it is very difficult to scale. Metalearning algorithms can consume enormous amounts of time and compute due to their nested optimization, and tend to grow complex because most deep learning frameworks are not well suited for them. This places an especially high incentive on developing and debugging metalearning algorithms on small-scale datasets such as MNIST-1D. 

We implemented a metalearning optimization for an MLP classifier on MNIST-1D with an explicitly written inner optimization loop using SGD. The gradient-based hyperparameter optimization converges to the optimal learning rate to be 0.62 regardless of whether the initial learning rate is too high or too low (Figure~\ref{fig:metalearn_lr}). The whole optimization process took only one minute on a CPU.

\subsection{Metalearning an activation function}

\begin{figure}[t]
    \centering
    \includegraphics[width=0.8\columnwidth]{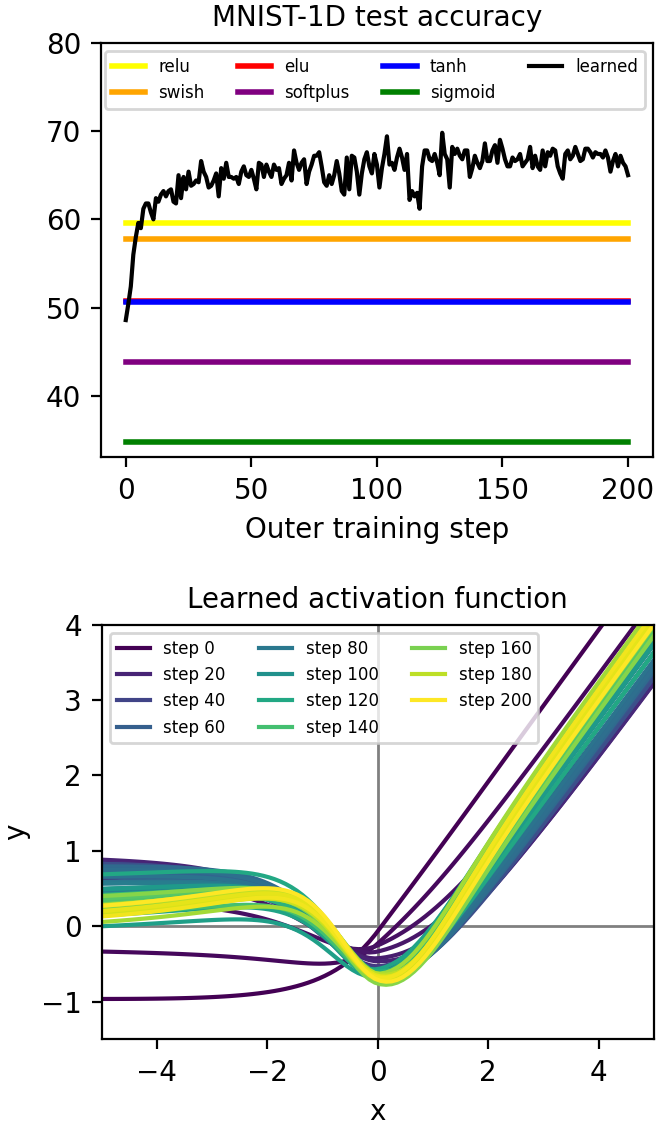}
    \caption{Metalearning an activation function. Starting from an ELU shape, we use gradient-based metalearning to find the optimal activation function for a neural network trained on the MNIST-1D dataset. The activation function itself is parameterized by a second (meta) neural network. Note that the ELU baseline (red) is obscured by the \texttt{tanh} baseline (blue) in the figure above. Runtime: $\sim$1 hour. \hfill\href{https://github.com/greydanus/mnist1d/blob/master/notebooks/metalearn-activation-function.ipynb}{\texttt{\textbf{[CODE]}}}}
    \label{fig:metalearn_afunc}
\end{figure}

The small size of MNIST-1D allows researchers to perform more challenging metalearning optimizations. For example, it permits the metalearning of an activation function --- something that to the best of our knowledge has not been studied before. We parameterized our classifier's activation function with a separate neural network (MLP with layer dimensionalities $1\to100\to100\to1$ using \texttt{tanh} activations, with outputs added to an ELU function such that it could be trained to produce perturbations to the ELU shape) and then learned its weights using meta-gradients. The learned activation function substantially outperformed common nonlinearities such as ReLU, Elu, and Swish (Figure~\ref{fig:metalearn_afunc}), achieving over 5 percentage points higher test accuracy. The resulting activation function had a non-monotonic shape with two local extrema (Figure~\ref{fig:metalearn_afunc}).

There has been work on optimizing activation functions \cite{clevert2015fast, ramachandran2017searching, vercellino2017hyperactivations}, but none has used analytical gradients computed via nested optimization. Moreover, some of these prior experiments \citep[e.g.][]{ramachandran2017searching} used multi-day training runs on large clusters of GPUs and TPUs, whereas our entire training took around 1 hour of CPU runtime.

\subsection{Self-supervised learning}

As shown in Table \ref{tab:main-benchmark}, logistic classification accuracy for MNIST-1D in pixel space was low (33\%). A powerful approach to self-supervised representation learning in computer vision is to rely on data augmentations: each input image is augmented twice, forming `positive pairs' which the network is trained to map to close locations in its output space while pushing away representations of other input images \citep{balestriero2023cookbook}. In particular, in SimCLR \citep{chen2020simple}, each positive pair is repulsed from all other positive pairs in the same mini-batch via the InfoNCE loss function.

\begin{figure}[t]
    \centering
    \includegraphics[width=\columnwidth]{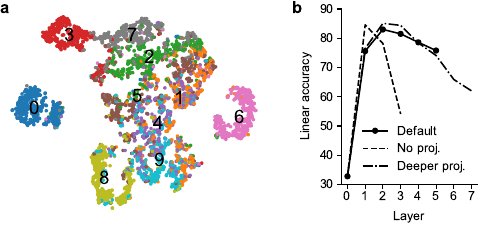}
    \caption{SimCLR-style \citep{chen2020simple} learning on MNIST-1D. \textbf{(a)} $t$-SNE embedding of the output representation after training ($n=5000$). \textbf{(b)} Linear classification accuracy after each layer. Layer 0 stands for the input (pixel space). Accuracy always peaks in the middle \citep{bordes2023guillotine}. CPU runtime: $\sim$5 minutes. \hfill\href{https://github.com/greydanus/mnist1d/blob/master/notebooks/self-supervised-learning.ipynb}{\texttt{\textbf{[CODE]}}}}
    \label{fig:ssl}
\end{figure}

We implemented the SimCLR algorithm for MNIST-1D, using a network with three convolutional and two fully-connected layers (`projection head') with output dimensionality $16$. Our data augmentations consisted of regressing out the linear slope, circularly shifting by up to 10 pixels, and then reintroducing a random linear slope. We achieved 82\% linear classification accuracy before the projection head in $\sim$1 minute of CPU training (for comparison, training SimCLR on CIFAR-10/100 datasets typically takes $\sim$10 GPU hours). In the output space, digits 0, 3, 6, and 8 appeared as isolated clusters (Figure~\ref{fig:ssl}a). Note that here we used both training and test sets of MNIST-1D for the self-supervised training, and the linear classifier was subsequently trained on the training set and evaluated on the test set.

Empirically, it has been observed that the representation quality (as measured via linear classification accuracy) is higher before the projection head rather than after \citep{chen2020simple}; removal of the projection head after training has been dubbed \textit{guillotine regularization} \citep{bordes2023guillotine} but remains poorly understood. We observed the same effect in our experiment: classification accuracy was the highest after the second layer (Figure~\ref{fig:ssl}b). Furthermore, when using a deeper projection head with four layers, or a network without any projection head at all, we achieved similar representation quality, and the accuracy always peaked in the middle (Figure~\ref{fig:ssl}b). This suggests that MNIST-1D is sufficiently rich to study cutting-edge open problems in self-supervised learning.

\subsection{Benchmarking pooling methods}

\begin{figure*}[t]
    \centering
    \includegraphics[width=\textwidth]{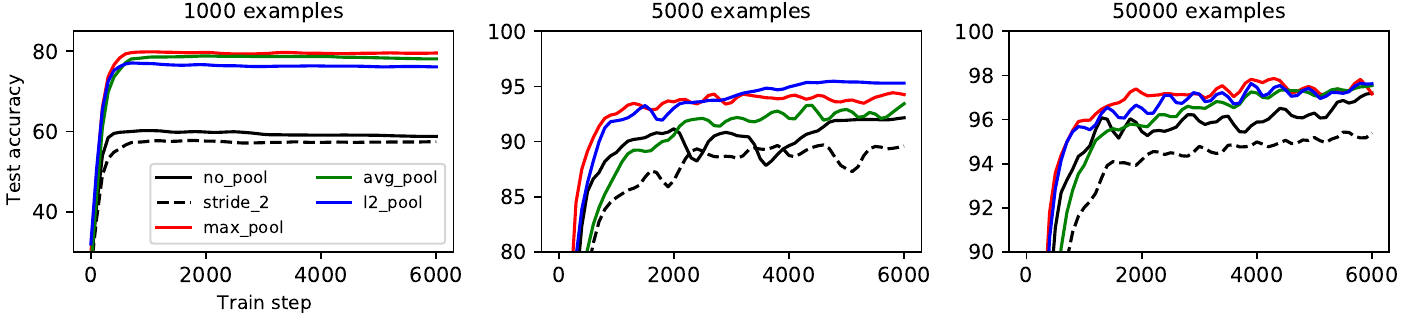}
    \caption{Benchmarking common pooling methods. Pooling was helpful in low-data regimes but hindered performance in high-data regimes. Runtime: $\sim$5 minutes. \hfill\href{https://github.com/greydanus/mnist1d/blob/master/notebooks/benchmark-pooling.ipynb}{\texttt{\textbf{[CODE]}}}}
    \label{fig:pooling}
\end{figure*}

In our final case study we asked: \textit{What is the relationship between pooling and sample efficiency?} Here we define pooling as any operation that combines activations from two or more neurons into a single feature. We are not aware of any prior literature on whether pooling makes models more or less sample efficient.

With this in mind, we trained CNN models for MNIST-1D classification with different pooling methods and training set sizes. Note that here we make use of the procedural generation of MNIST-1D that allows one to generate additional samples at will. We found that, while pooling (but not striding!) was very effective in low-data regimes, it did not make much of a difference when more training data was available (Figure~\ref{fig:pooling}). We hypothesize that pooling is a poor-man architectural prior which is better than nothing with insufficient data but restricts model expression otherwise.

\section{Discussion}

\paragraph{When to scale.} This paper is not an argument against large-scale machine learning research. That research has proven its worth and has come to represent one of the most exciting aspects of the ML research ecosystem. Rather, we wish to \emph{promote} small-scale machine learning research. Neural networks do not have problems with scaling or performance --- but they do have problems with interpretability, reproducibility, and training speed. We see carefully-controlled, small-scale experiments as a great way to address these problems.

In fact, small-scale research is complimentary to large-scale research. As in biology, where fruit fly genetics helped guide the Human Genome Project, we believe that small-scale research should always have an eye on how to successfully scale. For example, several of the findings reported in this paper are at the point where they could be investigated at scale. It would be interesting to show that large-scale lottery tickets also learn spatial inductive biases and feature local connectivity. It would also be interesting to try metalearning an activation function on a larger model in order to find an activation that can outperform ReLU and Swish in practical deep learning systems.

\paragraph{Understanding vs. performance.} There has been some debate over the relative value of understanding neural nets versus increasing their performance. Some researchers contend that a high-performing algorithm need not be interpretable as long as it saves lives or produces economic value. Others argue that hard-to-interpret deep learning models should not be deployed in sensitive real-world contexts until we understand them better. Both arguments have merit. However, we believe that the process of identifying things we do not understand about large-scale neural networks, reproducing them in toy settings like MNIST-1D, and then performing careful ablation studies to isolate their causal mechanisms is likely to improve both performance and interpretability in the long run.

\paragraph{Reducing environmental impact.} There is hope that deep learning will have positive environmental applications \cite{loehle1987applying, rolnick2019tackling}. This may be true in the long run, but so far, artificial intelligence has done little to solve environmental problems. Deep learning models do, however, require massive amounts of electricity to train and deploy \cite{strubell2019energy}. Running experiments on smaller datasets --- and waiting to scale until one has a solid grasp of the phenomena involved --- is a good way to reduce the electricity costs and environmental impact of this research.

\paragraph{The scaling down manifesto.} We would like to provocatively suggest that in order to explore the limits of how large we can scale neural networks, we may need to explore the limits of how small we can scale them first. Scaling models and datasets down in a way that preserves the nuances of their behaviors will allow researchers to iterate more quickly on fundamental and creative ideas. This fast iteration cycle is the best way to obtain insights on how to incorporate progressively more complex inductive biases into our models. We can then transfer these inductive biases across scales in order to dramatically improve the sample efficiency and generalization of large models. The MNIST-1D dataset is a first step in that direction.

\newpage

%% file: appendix.tex
\clearpage
\onecolumn
\section{Supplementary Figures}

\vspace{15em}
\begin{figure}[h!]
    \centering
    \includegraphics[width=0.35\textwidth]{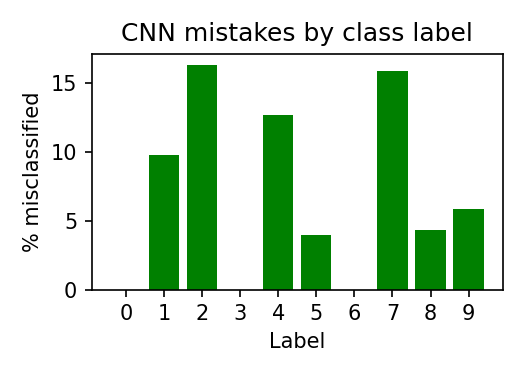}
    \hspace{2em}
    \includegraphics[width=0.35\textwidth]{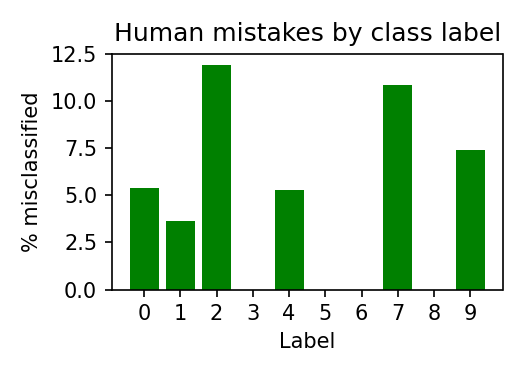}
    \caption{Classwise errors of a CNN and a human subject on the test split of the MNIST-1D dataset.
    As described in the main text, we estimated human performance on MNIST-1D by training one of the authors to perform classification and then evaluating his accuracy on 500 test images. His accuracy was 96\%. The CNN's accuracy was 94\%. Both humans and the CNN struggled primarily with classifying 2's and 7's, and to a lesser degree 4's. The human subject had a harder time classifying 9's whereas the CNN had a harder time classifying 1's. Both had zero errors classifying 3's and 6's. 
    It is interesting that a human could outperform a CNN on this task. Part of the reason may be that the CNN was only given 4000 training examples --- with more examples it could possibly match and eventually exceed the human baseline. Even though the data is low-dimensional, the classification objective is quite difficult and spatial/relational priors matter a lot. It may be that the architecture of the CNN prevents it from learning all of the tricks that humans are capable of using. 
    It is worth noting that modern CNNs can outperform human subjects on most large-scale image classification tasks like ImageNet. But in our tiny benchmark  a human was still competitive.
    }
    \label{fig:distribution}
\end{figure}

\begin{figure*}[h!]
    \centering
    \includegraphics[width=\textwidth]{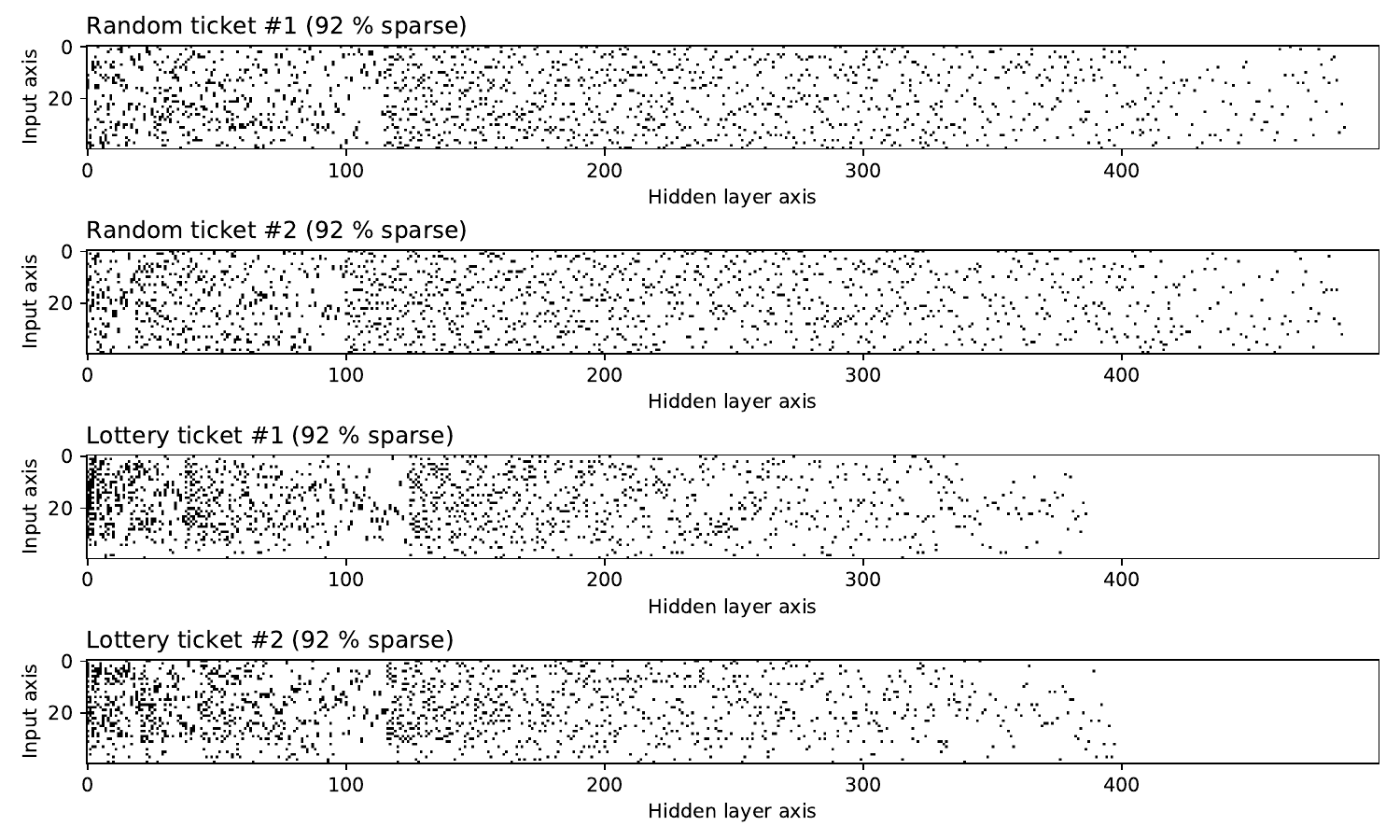}
    \vspace{0.1pt}
    \caption{First layer weight masks of random tickets and lottery tickets. We sorted the masks along their hidden layer axes, according to the number of adjacent unmasked parameters. This helps to reveal a bias towards local connectivity in the lottery ticket masks. Notice how there are many more vertically-adjacent unmasked parameters in the lottery ticket masks. These vertically-adjacent parameters correspond to local connectivity along the input dimension, which in turn biases the sparse model towards data with spatial structure.}
    \label{fig:lottery_masks}
\end{figure*}

